\documentclass[11pt]{article}

\usepackage[preprint]{acl}

\usepackage{times}
\usepackage{latexsym}

\usepackage[T1]{fontenc}

\usepackage[utf8]{inputenc}

\usepackage{microtype}

\usepackage{inconsolata}

\usepackage{graphicx}

\usepackage[dvipsnames]{xcolor}
\usepackage{booktabs}
\usepackage{multirow}
\usepackage{dirtree}
\usepackage{amsmath}
\usepackage{tcolorbox}

\usepackage{caption}
\tcbuselibrary{breakable,skins}
\usepackage{enumitem}

\definecolor{promptbg}{RGB}{245,245,250}
\definecolor{promptframe}{RGB}{70,70,160}

\usepackage{enumitem}
\setlist[itemize]{
    noitemsep,
    leftmargin=*,
}
\setlist[enumerate]{
    noitemsep,
    topsep=0pt,
}

\title{LexIssue: Benchmarking Legal Issue Identification in Chinese Civil Litigation}

\author{
  \textbf{Huiyuan Xie\textsuperscript{1}\thanks{Equal contribution.}}, 
  \textbf{Yuqin Huang\textsuperscript{1}\footnotemark[1]}, 
  \textbf{Zhicheng Hao\textsuperscript{1}}, 
  \textbf{Yida Cai\textsuperscript{1,2}},
  \textbf{Shaochun Wang\textsuperscript{3}},\\
  \textbf{Zhenghao Liu\textsuperscript{4},}
  \textbf{Yuxiao Ye\textsuperscript{1}\thanks{Corresponding author.}}\\
  \textsuperscript{1}Tsinghua University
  \textsuperscript{2}Peking University
  \textsuperscript{3}Modelbest Inc.
  \textsuperscript{4}Northeastern University\\
  \texttt{\{xieh,yeyuxiao\}@tsinghua.edu.cn~~huang-yq23@tsinghua.org.cn}
}

\begin{document}
\maketitle
\begin{abstract}
Identifying the issues disputed between litigating parties is a crucial component of real-world litigation. However, legal issues remain comparatively underexplored in legal AI research. In this work, we study the computational modelling of legal issue identification in litigation. We introduce a legally grounded hierarchical schema that represents legal issues through both free-form issue descriptions and structured legal categories, and formulate legal issue identification as two complementary tasks: legal issue generation and legal issue classification. Based on this formulation, we construct LexIssue, a benchmark containing 430 real-world Chinese civil litigation cases and 1,303 expert-annotated disputed legal issues. We further develop an issue-centric legal knowledge base spanning 27 causes of action and 441 candidate legal issue entries to support retrieval-augmented reasoning. Experimental results across a diverse set of models show that retrieval-augmented generation using the constructed legal issue knowledge base consistently improves performance in identifying disputed legal issues and their corresponding legal attributes.\footnote{Data and code are available at \url{https://github.com/thunlp/LexIssue}.}
\end{abstract}

\section{Introduction}\label{sec:intro}

\begin{figure*}
    \centering
    \includegraphics[width=\linewidth]{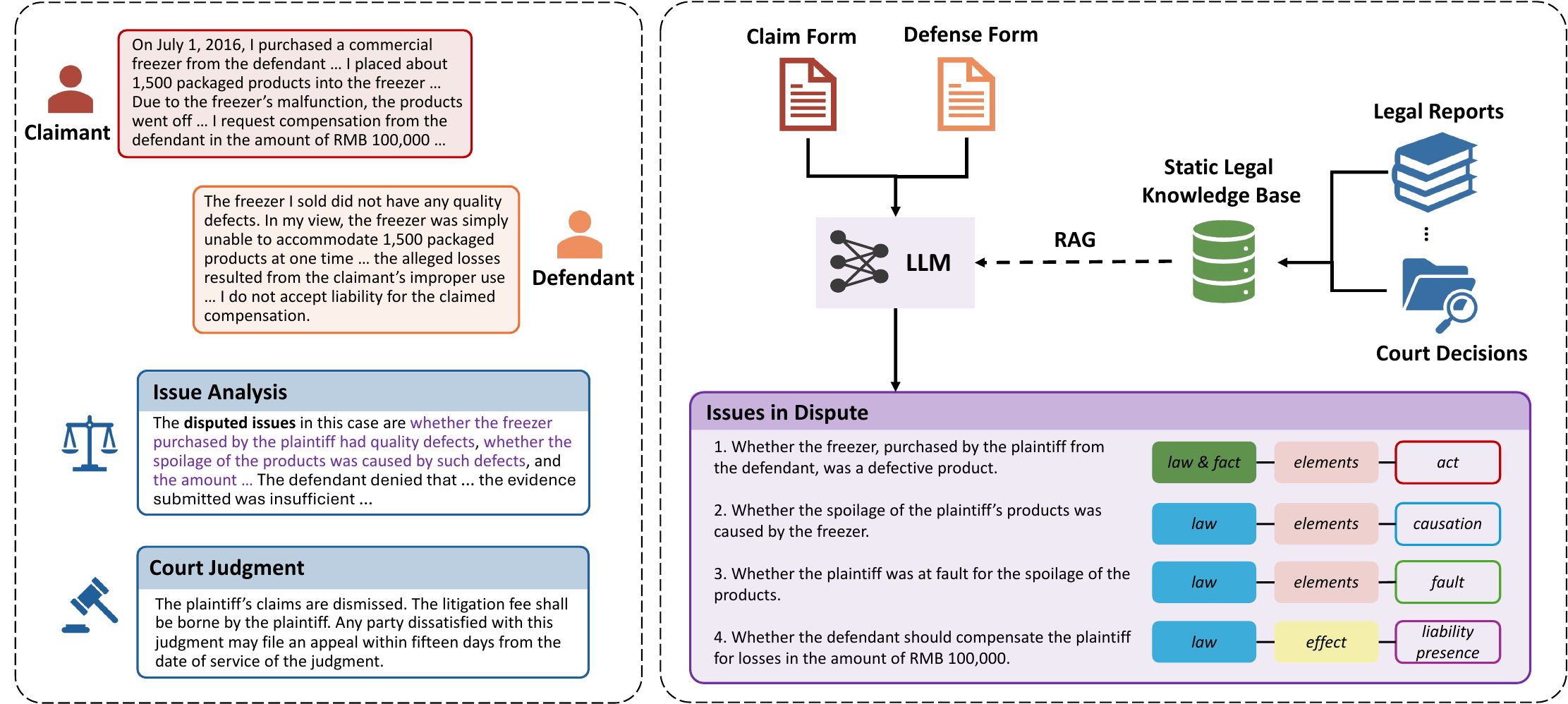}
    \caption{\textbf{Overview of the legal issue identification process}. \textbf{Left}: An illustrative product liability dispute in which the claimant alleges that a freezer purchased from the defendant is defective and seeks compensation for losses, while the defendant denies the existence of defects, disputes causation and fault, and refuses compensation. The issue analysis content is extracted from the judge's reasoning section in the original court judgment. \textbf{Right}: An illustration of the legal issue identification task, where models identify disputed legal issues from the litigated case and generate both descriptive issue summaries and structured legal category labels. In the illustrated three-tier legal category hierarchy, the first-tier labels \textit{law \& fact} and \textit{law} correspond to \textit{mixed issues of law and fact} and \textit{substantive issues of law}, respectively. The second-tier labels \textit{elements} and \textit{effect} correspond to \textit{elements of a claim} and \textit{legal effect}. The third-tier labels \textit{act}, \textit{causation}, \textit{fault}, and \textit{liability presence} correspond to the schema categories \textit{tortious act}, \textit{causation}, \textit{fault}, and \textit{presence or absence of liability}, respectively.}
    \label{fig:overview}
\end{figure*}

In legal practice, legal issues typically refer to the specific points of dispute arising from the parties' claims that must be resolved in order to determine a case. 
Legal adjudication is fundamentally organised around disputed legal issues~\cite{fuller1978forms}. In litigation practice, judges typically structure legal analysis around the key issues contested between parties: the particulars that the parties disagree on, which legal questions must be resolved, and how those questions should be determined through the application of law (see Figure~\ref{fig:overview} for an illustration). A court cannot meaningfully apply legal rules before determining what is actually disputed in the case. This issue-centred structure is also deeply embedded in legal education and legal writing. Widely used analytical frameworks such as IRAC (abbreviation for \textit{Issue}, \textit{Rule}, \textit{Application}, \textit{Conclusion})~\cite{metzler2002importance,turner2012finding} treat issue identification as the foundational step of adjudicative reasoning. 

Despite this centrality, legal issues remain comparatively underexplored in legal AI research. Existing work has made substantial progress on various legal AI tasks~\cite{xiao2018cail,cui2023survey,xie2024clc_uket,ribeiro2025information,su2025judge}. However, comparatively little attention has been paid to modelling legal issues themselves as explicit computational objects. This gap is particularly important as identifying what is being disputed is a prerequisite for many realistic legal reasoning and dispute-resolution workflows.

The task of identifying legal issues in disputes is both conceptually and operationally challenging. In judicial decisions, disputed issues often shape legal reasoning throughout the judgment, yet they are not always explicitly enumerated or formally stated. Even when judges articulate the issues, they are typically expressed through highly free-form legal language. Semantically similar issues may differ substantially in framing and granularity. Moreover, issues may be merged, decomposed, or implicitly embedded within broader judicial reasoning structures. Legal issue identification therefore requires more than surface-level text extraction: it requires modelling the legal meaning and doctrinal function of disputed issues within litigation.
 
Existing computational work on legal issue identification remains limited. Prior studies typically focus on a narrow range of civil causes of action~\cite{cail2024} and often formulate issue identification as multiple-choice classification tasks where models select dispute categories from predefined options~\cite{fei2024lawbench,dai2025laiw}, or adopt a generative formulation but evaluate issue descriptions using generic natural language generation (NLG) metrics such as ROUGE~\cite{lin2004rouge}. Such metrics rely heavily on lexical overlap and are poorly suited to legal issue identification, where legally equivalent issues may vary substantially in framing and style.


In this work, we examine the computational modelling of legal issue identification in litigation. Our contributions are summarised as follows:

\begin{itemize}
    \item We introduce a legally-grounded, hierarchical schema that models legal issues through both descriptive representations (free-form legal issue summaries) and structured categorical representations (legal attribute categories).
    \item Based on this schema, we formulate the task of \textit{legal issue identification}, consisting of two complementary components: (1) \textit{legal issue generation}, which requires models to generate free-form issue descriptions, and (2) \textit{legal issue classification}, which requires models to predict legal categories capturing the nature and attributes of the issues. This formulation enables the evaluation of both generative capabilities and legally-bounded reasoning about dispute structure.
    \item We construct LexIssue, a benchmark for \textit{legal issue identification} containing expert-annotated issue representations for real-world litigated cases. Each test instance consists of claim and defence statements from litigated parties, together with expert-annotated issues in dispute. The resulting benchmark contains 430 Chinese civil cases and 1,303 fine-grained annotations covering diverse aspects of disputed legal issues.
    \item We construct a legal issue knowledge base, providing issue-centric legal knowledge spanning 27 causes of action and 441 candidate legal issue items, and can serve as an external source for retrieval-augmented generation (RAG) systems.
    \item We evaluate a diverse set of models under zero-shot, legal prompting, and RAG settings. Results show that RAG consistently improves performance across models, whereas directly injecting legal issue knowledge through prompting alone is unstable and may even degrade performance.
\end{itemize} 

\section{Related Work}\label{sec:rw}

\subsection{Legal Reasoning and Legal Issues}

Legal reasoning has long been a central topic in jurisprudence and legal theory. Early formalist traditions often conceptualised adjudication as a process of mechanical rule application, in which judges derive conclusions deductively from legal rules and case facts. Under this classical view, legal reasoning is framed as a syllogism consisting of a major premise (legal rules), a minor premise (facts of a case), and a conclusion (the judgment). 

However, subsequent jurisprudential scholarship argued that this deductive framing is insufficient to capture the realities of legal reasoning. Legal rules are often indeterminate or open to competing interpretations, facts may themselves be disputed, and the relevance or characterisation of facts frequently depends on normative and doctrinal judgment. As a result, legal reasoning cannot be reduced to purely mechanical deduction from fixed premises. Instead, many scholars conceptualise legal reasoning as interpretative, argumentative, and adversarial in nature~\cite{levi1948introduction,holmes1897path,maccormick1994legal,sartor2009defeasibility}. 
In this view, legal reasoning operates through the identification and resolution of contested legal questions arising from disputed facts, implicitly positioning legal issues as an essential unit of legal analysis.

The important role of legal issues is also reflected in legal education and litigation practice. 
Legal professionals are extensively trained to identify legally material disputes, distinguish relevant from irrelevant facts, and formulate legal questions that structure judicial analysis. This tradition is commonly operationalised through the IRAC (Issue, Rule, Application, Conclusion) framework and its variants~\cite{metzler2002importance,turner2012finding}, which treat issue identification as the organising structure through which legal rules and factual analysis are connected. Under these frameworks, identifying the disputed legal issues is not merely a preliminary drafting exercise, but a fundamental step that shapes the subsequent application of legal doctrine and reasoning process.

\subsection{Computational Modelling of Legal Issues}

Despite the theoretical and practical importance of legal issues in jurisprudence, legal education, and litigation practice, the computational modelling of legal issues remains underexplored. 

One closely related line of research concerns computational legal reasoning. Early approaches largely relied on symbolic and rule-based systems, conceptualising legal reasoning through explicit rule representation and deductive inference~\cite{gardner1987artificial}. Subsequent work shifted towards emphasising interpretative and argumentative views of legal reasoning. For example, \citet{ashley1991modeling} conceptualised legal reasoning as fundamentally adversarial, factor-based, and issue-framed, where adjudication operates through structured disputes concerning legally salient factors and competing interpretations of cases. Related work on argumentation-based legal reasoning~\cite{benchcapon2009argumentation} further modelled adjudication as structured argumentative conflict governed by competing claims, rebuttals, and justificatory relations. Although these studies closely engage with the concept of legal issues, issues are typically embedded within broader theories of legal reasoning, rather than treated as standalone computational representations requiring explicit treatment.


More recently, several legal NLP benchmarks~\cite{fei2024lawbench,cail2024,dai2025laiw} have included tasks related to dispute focus identification or issue extraction. However, these tasks are often formulated as narrow classification problems or evaluated using generic natural language generation (NLG) metrics such as ROUGE. Such formulations simplify the open-ended and legally-interpretative nature of issue identification in realistic litigation settings, where semantically equivalent legal issues may vary substantially in wording, framing, and granularity. In contrast, our work treats legal issues as both descriptive and structured legal representations, and examines legal issue identification as a distinct computational task grounded in dispute-centred legal reasoning.


\section{Schema for Representing Legal Issues}\label{sec:schema}

In judicial practice, issues in dispute are typically expressed in highly free-form natural language, rather than through unified or standardised representations. As legal pleadings and judgments are primarily drafted for adjudicative rather than computational purposes, issue formulations often reflect stylistic preferences or case-specific reasoning structures. While free-form issue descriptions preserve the substantive legal meaning of disputes, their lack of structure makes reliable annotation, evaluation, and large-scale computational modelling difficult. A structured schema is therefore necessary to bridge natural legal expression and computationally tractable legal abstraction.

To address this challenge, we introduce a dual-layer schema for representing legal issues, comprising two complementary layers: a descriptive summary layer and a structured legal category layer. 

The first layer contains case-specific free-form issue summaries. This layer preserves the richness of natural language and contextuality of issue formulations as they appear in real litigation. The second layer assigns each issue to a tiered legal category hierarchy capturing its doctrinal nature and reasoning role. At the first tier, the schema distinguishes among: (1) \textit{substantive issues of law}, (2) \textit{substantive issues of fact}, (3) \textit{mixed issues of law and fact}, and (4) \textit{procedural issues}. This distinction reflects the observation that litigation disputes may concern legal interpretation, factual determination, the application of legal standards to contested facts, or procedural questions. 

For \textit{substantive issues of law} and \textit{mixed issues of law and fact}, the schema further introduces second- and third-tier categories. At the second tier, issues are classified into \textit{elements of a claim}, \textit{legal effects}, and multiple categories of \textit{defences}. These categories correspond to distinct reasoning functions within adjudication. For example, claim-element issues concern whether liability requirements such as causation or fault are satisfied, legal-effect issues concern the consequences of established liability, and defence-related issues concern whether a right is prevented, extinguished, or temporarily barred. A full illustration of the schema is provided in Appendix~\ref{app:schema}.

Explicitly modelling these distinctions enables the benchmark to evaluate whether models understand the legal nature and reasoning role of disputed issues, rather than merely extracting semantically related phrases. The schema additionally improves annotation consistency and evaluation reliability by providing a stable legal abstraction over heterogeneous issue formulations.


\section{Legal Issue Identification}\label{sec:task_and_benchmark}

Based on the schema, we formulate the task of \textit{legal issue identification} (see Figure~\ref{fig:overview} for an illustration), and construct the LexIssue benchmark, an evaluation dataset for legal issue identification with expert-annotated legal issue representations.

\subsection{The Task of Legal Issue Identification}\label{subsec:lii_task}

Legal issue identification requires a model to correctly identify issues in dispute and their corresponding legal attributes. The schema designed in Section~\ref{sec:schema} provides a principled guidance for formalising this task. More specifically, given the claims from claimants and defences from defendants of a case, the legal issue identification task examines two complementary components: (1) \textit{legal issue generation}, which requires models to generate free-form descriptions of disputed issues, and (2) \textit{legal issue classification}, which requires models to predict legally grounded categories capturing the nature and attributes of the issues. This formulation enables the evaluation of both generative capabilities and legally bounded reasoning concerning issue nature and functionality in disputes. 

\subsection{The LexIssue Benchmark}\label{subsec:lexissue_benchmark}

We construct the LexIssue benchmark to model legal issue identification using real-world litigated civil cases. Civil litigation provides a particularly suitable setting for studying legal issue identification because civil adjudication is fundamentally dispute-centred: courts typically organise judicial reasoning around contested claims, defences, and disputed legal questions raised by the parties. 

\subsubsection{Data Construction}

We begin by randomly sampling 450 first-instance Chinese civil cases collected from~\citet{cjo2013}. Each court judgment contains multiple components of the litigation process, such as party statements, case facts, evidentiary analysis, judicial reasoning, and final judgments.

The benchmark construction consists of two major steps: (1) synthesising litigation-stage inputs, and (2) expert annotation of legal issues.

\paragraph{Synthesising claim and defence forms.} In real litigation practice, legal disputes are typically framed through claim forms and defence forms submitted by the parties. A claim form typically contains the claimant's claims, factual assertions, legal arguments, and requested remedies, while a defence form contains the defendant's responses, counterarguments, factual disputes, and legal defences. However, original litigation pleadings are difficult to obtain at scale in publicly accessible data. To approximate realistic litigation-stage inputs, we use GPT-4o to reconstruct synthetic claim and defence forms from judges' recounting of the parties pleadings in court judgments.

\paragraph{Expert annotation of legal issues.} To ensure the quality of the reference legal issues, we conduct rigorous expert annotation under the legal issue schema introduced in Section~\ref{sec:schema}. Legal experts with postgraduate-level legal training manually review the litigated cases and annotate the disputed legal issues together with their corresponding legal attribute categories. The annotation process is conducted under the supervision of a senior researcher with extensive experience in legal AI and legal annotation workflows. Further details regarding the annotation process are provided in Appendix~\ref{app:human_annotation}.

Among the 450 prepared cases, 20 cases are reserved for annotation standard alignment and for the development and refinement of the evaluation protocol described in Section~\ref{subsubsec:evaluation}. After annotation, the resulting evaluation benchmark contains 430 real-world Chinese litigation cases spanning 10 distinct causes of action categories, and 1,303 expert-annotated issues in dispute, each associated with a descriptive issue summary and structured legal category labels. On average, each case contains 3.03 disputed legal issues, with the maximum number of issues in a single case being 7. In terms of top-level legal attribute distribution, the majority (80.51\%) of annotated issues fall under the \textit{substantive issues of law} category, and 11.36\% of the issues are categorised as \textit{substantive issues of fact}, while \textit{mixed issues of law and fact} and \textit{procedural issues} account for 6.29\% and 1.84\% of the annotations, respectively.

\subsubsection{Evaluation Design}\label{subsubsec:evaluation}

Evaluating legal issue identification is challenging as dispute issues are expressed in highly free-form language. Semantically equivalent issues may differ substantially in wording and abstraction level, making lexical-overlap-based metrics such as ROUGE unreliable for this task (see Section~\ref{sec:eval_analysis} for empirical analysis). The core challenge therefore lies in establishing reliable alignments between model-predicted issues and expert-annotated reference issues before metric computation.

\paragraph{Prediction-reference alignment.} To address this problem, we design a two-stage, LLM-assisted prediction-reference alignment framework consisting of: (1) \textit{candidate similarity mapping} and (2) \textit{issue-level alignment verification}.

In the first stage, the framework performs coarse-grained semantic correspondence assessment. For each predicted issue, the LLM judge retrieves the single most semantically similar reference from the set of gold-reference issues for the case. This stage functions as a semantic retrieval step, identifying potentially corresponding prediction–reference issue pairs without enforcing strict equivalence.

In the second stage, the framework performs stricter issue-level alignment verification. The candidate prediction–reference pairs identified in the first stage are evaluated by an LLM judge using a rubric-based matching criterion designed by legal experts (see prompt in Appendix~\ref{app:rubric_prompt}). The objective of this stage is to determine whether the prediction and reference issue genuinely refer to the same underlying disputed legal issue, rather than merely exhibiting broad semantic similarity. Pairs receiving scores above a predefined acceptance threshold are retained as successfully aligned prediction-reference pairs. The matching threshold is verified by legal experts both during the rubric design process and through empirical inspection of the resulting filtering behaviour to ensure that the retained alignments reflect legally meaningful issue equivalence.

The two-stage decomposition is designed to improve the reliability of LLM-based alignment judgment by separating coarse semantic retrieval from strict, legally-informed equivalence verification. A detailed description of the issue alignment procedure is provided in Appendix~\ref{app:evaluation_design}.

We employ Qwen-3.5-27B~\cite{qwen3.5} as the LLM judge in both stages. The model is selected based on validation experiments conducted on 103 expert-annotated issue instances from 20 validation cases, where it demonstrates the highest agreement with expert human judgments compared to conventional NLG metrics~\cite{lin2004rouge,papineni2002bleu,popovic2015chrf} and embedding-based semantic similarity methods~\cite{li2023towards,xiao2023bge_embedding,muennighoff2022sgpt,kusupati2024matryoshka,reimers2019sentence_bert,youdao2023bcembedding,ming2022text2vec}. The use of an open-weight model additionally enables local deployment and reproducible evaluation.

\paragraph{Evaluation metrics.} Evaluation is performed based on the final prediction-reference alignments obtained from the above steps. 

\textbf{Legal Issue Generation (LIG).} For each test case, case-level F1 is computed based on the successfully aligned prediction-reference pairs. Final performance on LIG is reported as the average case-level Weighted-F1 over the test set.

\textbf{Legal Issue Classification (LIC).} For each test case, we first use the issue alignment pipeline to determine which predicted issues are successfully aligned with gold-reference issues. A predicted issue category is counted as a true positive only if the issue is successfully aligned with a reference issue and its predicted legal category matches the reference label under the corresponding evaluation granularity. We consider three levels of evaluation granularity:

\begin{itemize}
\item \textbf{Tier-1.} This setting evaluates only the first tier of the issue category structure and therefore represents the loosest evaluation criterion.







\item \textbf{Tier-2.} This setting evaluates similarity at the first two issue category tiers. 


\item \textbf{Tier-3.} This setting treats the complete hierarchical category path as a single label and therefore represents the strictest evaluation criterion.





\end{itemize}

Final LIC performance under each granularity setting is reported as the average case-level Weighted-F1 over the test set.

\section{Legal Issue Knowledge Base}\label{sec:knowledge_base}

Legal issue identification requires recognising the underlying legal questions embedded within disputes. In litigation practice, disputes within the same \textit{cause of action} often revolve around relatively stable and recurring issue patterns. For example, tort disputes frequently involve issues concerning causation, fault, or damages, while contract disputes commonly concern contract formation, performance, or breach. These recurring issue structures form an important part of legal practitioners' doctrinal and experiential knowledge. Motivated by this observation, we construct a legal issue knowledge base that organises commonly occurring disputed legal questions into an issue-centric legal repository according to causes of action.

The legal issue knowledge base is curated by legal experts from two complementary sources: (1) doctrinal and practical legal materials, and (2) real-world litigated cases.

\paragraph{Knowledge derived from legal reports and judicial materials.} The first source consists of legal essays, judicial reports, explanatory legal publications, and practical guidance documents produced by courts and law firms. From these materials, legal experts first identify discussions related to disputed legal issues and organise them according to causes of action. The extracted issue descriptions are then rewritten into unified issue-summary formulations. This process produces candidate issue summaries representing doctrinally important and practically recurring disputed issues.

\paragraph{Knowledge derived from real-world litigated cases.} The second source consists of real-world judicial decisions spanning different causes of action. For each cause of action, we randomly sample 100 judicial decisions. Legal experts then review the judicial reasoning sections of these cases and extract issue-related statements discussed during adjudication. These issue discussions are subsequently summarised into concise and reusable legal issue descriptions.

Compared with doctrinal and explanatory legal materials, litigated cases provide issue formulations grounded in realistic judicial reasoning and dispute resolution contexts. This complements the more abstract and doctrinally oriented issue formulations extracted from legal essays and practical legal guidance materials.

\paragraph{Knowledge base integration.} The extracted issue summaries from both sources are subsequently merged into a unified issue-centric knowledge base. During integration, legal experts perform deduplication, granularity normalisation, and stylistic refinement to improve consistency across issue formulations. Semantically overlapping issues are consolidated, while overly broad or excessively specific issue descriptions are reorganised into more coherent levels of abstraction.

The resulting knowledge base provides issue-centric legal knowledge spanning 27 Chinese civil causes of action and 441 candidate legal issue entries, and can further serve as an external knowledge source for retrieval-augmented generation systems and other downstream legal reasoning tasks.

It is important to distinguish the knowledge base entries from the case-specific issue descriptions in our schema. The descriptive layer of the schema described in Section~\ref{sec:schema} captures the actual issues disputed in an individual litigated case. By contrast, the knowledge base contains generic issue summaries describing potentially applicable legal issues commonly arising within a category of disputes. These issue summaries therefore function as reusable doctrinal abstractions rather than case-specific annotations.


\section{Experiments and Results}\label{sec:experiments_and_results}

\begin{table}[t!]
    \centering
    \small
    \setlength{\tabcolsep}{4pt}
    \resizebox{0.85\linewidth}{!}{
    \begin{tabular}{llccc}
    \toprule
    \multirow{2}{*}{\textbf{Model}} & \multirow{2}{*}{\textbf{LIG}} & \multicolumn{3}{c}{\textbf{LIC}} \\
    \cmidrule(lr){3-5} & & \textbf{Tier-1} & \textbf{Tier-2} & \textbf{Tier-3} \\
    \midrule
    \textbf{GPT-5.2} &0.542& 0.346& 0.255& 0.185 \\ 	 	 	 
    \quad w/ LP & 0.475& 0.279& 0.191& 0.133\\
    \quad w/ RAG & 0.634& 0.460& 0.335& 0.243\\\addlinespace
    \textbf{Gemini-3-Flash} &0.701& 0.581& 0.459& \textbf{0.357}\\
    \quad w/ LP & 0.699& 0.544& 0.394& 0.292\\
    \quad w/ RAG & \textbf{0.715}& \textbf{0.597}& \textbf{0.467}& 0.339\\
    \midrule
    \textbf{DeepSeek-R1} & 0.538& 0.366& 0.282& 0.182\\
    \quad w/ LP & 0.502& 0.267& 0.176& 0.118\\
    \quad w/ RAG & 0.594& 0.444& 0.365& 0.267\\\addlinespace
    \textbf{DeepSeek-V4-Flash} &  0.583& 0.366& 0.285& 0.204\\
    \quad w/ LP & 0.571& 0.343& 0.257& 0.180\\
    \quad w/ RAG & 0.628& 0.477& 0.402& 0.295\\\addlinespace
    \textbf{Qwen-3-8B} & 0.464& 0.274& 0.183& 0.098\\
    \quad w/ LP & 0.431& 0.230& 0.168& 0.085\\
    \quad w/ RAG & 0.495& 0.301& 0.249& 0.131\\\addlinespace
    \textbf{Llama-3.1-8B} & 0.450& 0.252& 0.204& 0.101\\
    \quad w/ LP & 0.445& 0.306& 0.224& 0.109\\
    \quad w/ RAG & 0.457& 0.271& 0.210& 0.097\\
    \midrule
    \textbf{LegalOne-R1} & 0.417& 0.291& 0.184& 0.087\\
    \quad w/ LP & 0.403& 0.274& 0.198& 0.079\\
    \quad w/ RAG & 0.424& 0.357& 0.253& 0.138\\\addlinespace
    \textbf{DISC-LawLLM} & 0.389& 0.324& 0.079& 0.042\\
    \quad w/ LP & 0.380& 0.292& 0.153& 0.076\\
    \quad w/ RAG & 0.400& 0.286& 0.040& 0.021\\
    \bottomrule
    \end{tabular}
    }
    \caption{Evaluation results on the LexIssue benchmark. LIG denotes the \textit{legal issue generation} task, whilst LIC denotes the \textit{legal issue classification} task. Entries labelled with the model name alone (e.g., GPT-5.2) correspond to the zero-shot inference setting. The variant ``w/ LP'' denotes models evaluated under the legal prompting setting, while ``w/ RAG'' denotes models augmented with RAG using the external legal issue knowledge base.}
    \label{tab:main_results}
\end{table}

\subsection{Baselines}

We evaluate a collection of large language models (LLMs) on LexIssue, including general-purpose closed-source LLMs such as GPT-5.2~\cite{openai2025gpt5_2} and Gemini-3-Flash~\cite{google2025gemini3}, general-purpose open-source LLMs including DeepSeek-R1\footnote{Although DeepSeek-R1 and DeepSeek-V4-Flash are open-source models, we experimented via API calls due to computational resource considerations.}~\cite{guo2025deepseek_r1}, DeepSeek-V4-Flash~\cite{deepseekv4}, Qwen-3-8B~\cite{yang2025qwen_3}, and Llama-3.1-8B~\cite{grattafiori2024llama_3.1}, and legal-domain LLMs such as LegalOne-R1~\cite{li2026legalone} and DISC-LawLLM~\cite{yue2023disc}. We experiment with three inference settings (see implementation details in Appendix~\ref{app:experiment_setup}):
 
\textbf{Zero-shot.} Models are prompted using task instructions only, without demonstrations, reasoning guidance, or external knowledge augmentation.

\textbf{Legal prompting.} Models are provided with additional guidance, including definitions of disputed legal issues, explanations of schema labels, and litigation-oriented chain-of-thought instructions.

\textbf{RAG-enhanced.} In the retrieval-augmented generation (RAG) setting, models are augmented with the external legal issue knowledge base introduced in Section~\ref{sec:knowledge_base}. For each case, candidate issue entries associated with the same cause of action are retrieved and reranked using Qwen3-Reranker-8B~\cite{qwen3embedding} based on their relevance to the claims and defences. The top-5 retrieved issue entries are injected into the prompt as external legal references for generation.

\subsection{Evaluation Results on LexIssue}

We evaluate widely used LLMs on the LexIssue benchmark under zero-shot, legal prompting (LP), and RAG-enhanced settings. 

\paragraph{Zero-shot.} As shown in Table~\ref{tab:main_results}, among all evaluated LLMs, Gemini-3-Flash achieves the strongest zero-shot performance across both LIG and LIC tasks, reaching $0.701$ on LIG and $0.581$, $0.459$, and $0.357$ on Tier-1, Tier-2, and Tier-3 for LIC.
Among open-source LLMs, DeepSeek-V4-Flash achieves the strongest overall performance, even surpassing GPT-5.2 on multiple metrics. In contrast, smaller LLMs such as Qwen-3-8B and Llama-3.1-8B perform relatively poorly on both tasks, indicating that smaller models still require substantial improvement for legal issue recognition and reasoning.

Interestingly, under comparable parameter scales, legal-domain-specific LLMs even underperform general open-source LLMs. Specifically, both legal-domain-specific LLMs perform comparably to or substantially worse than Qwen-3-8B and Llama-3.1-8B on most metrics. Notably, DISC-LawLLM, despite being trained on legal corpora, achieves the worst performance among all evaluated LLMs except on LIC Tier-1.

\paragraph{Legal prompting.} Under the LP setting, which incorporates structured legal reasoning prompts, the results show that such prompting strategies do not consistently improve LLM performance. In many cases, legally informed prompting even leads to performance degradation. These observations suggest that directly injecting legal issue knowledge through prompting alone is unstable and may negatively affect models' ability to identify and structure disputed legal issues.

\paragraph{RAG-enhanced.} In contrast, incorporating external legal knowledge through RAG consistently improves performance across almost all models and evaluation metrics, with only a few exceptions, such as DISC-LawLLM on LIC, and Gemini-3-Flash on LIC Tier-3.
These results indicate that external legal knowledge is highly beneficial for legal issue identification, a knowledge-intensive task that involves structured legal concepts and domain-specific reasoning over litigation disputes.

\begin{table}[t!]
\centering
\small
\begin{tabular}{lccc}
\toprule
\textbf{Rater} & \textbf{Sim.} & \textbf{Align.} & \textbf{Agr.} \\
\midrule
Qwen-3.5-27B & \textbf{0.965} & \textbf{0.852} & \textbf{0.668} \\
ROUGE\_1 & 0.870 & 0.635 & 0.178 \\
gte-large-zh & 0.870 & 0.665 & 0.254 \\
\bottomrule
\end{tabular}
\caption{Comparison of representative instances for three categories of rating approaches: LLM-as-a-Judge (Qwen-3.5-27B), NLG-based metrics (ROUGE\_1), and embedding-based methods (gte-large-zh). \textit{Sim.} denotes the accuracy achieved in the similarity mapping step; \textit{Align.} denotes the accuracy achieved in the alignment verification step; \textit{Agr.} denotes the overall agreement with human judgments.}
\label{tab:rater_results}
\end{table}

\begin{table}[t!]
\centering
\small
\begin{tabular}{lccc}
\toprule
\textbf{Model} & \textbf{Tier-1} & \textbf{Tier-2} & \textbf{Tier-3} \\
\midrule
GPT-5.2           & \textbf{0.971} & 0.707 & 0.513 \\
Gemini-3-Flash    & 0.900 & \textbf{0.709} & \textbf{0.550} \\
DeepSeek-R1       & 0.877 & 0.653 & 0.421 \\
DeepSeek-V4-Flash & 0.851 & 0.657 & 0.467 \\
Qwen-3-8B         & 0.908 & 0.583 & 0.310 \\
Llama-3.1-8B      & 0.838 & 0.658 & 0.326 \\
LegalOne-R1       & 0.908 & 0.561 & 0.266 \\
DISC-LawLLM       & 0.932 & 0.220 & 0.119 \\
\bottomrule
\end{tabular}
\caption{
Model performance on legal issue classification under Tier-1, Tier-2, and Tier-3 settings, calculated only on the subset of successfully aligned issues.
}
\label{tab:lic_results}
\end{table}

\section{Discussion on Evaluation Design}\label{sec:eval_analysis}

A central challenge of legal issue identification lies in evaluation design, which requires aligning free-form predicted issue descriptions with gold-reference issues before assessing issue quality and label correctness. To investigate this challenge, we compare three types of similarity assessment approaches: (1) NLG-based metrics, (2) embedding-based similarity methods, and (3) LLM-as-a-Judge evaluation, on a held-out set of 230 expert-annotated issue instances. We report their accuracy across the two stages of the issue alignment pipeline and their overall agreement with human judgments in Table~\ref{tab:rater_results}, providing insights into the reliability of different rating approaches and difficulty of evaluating legal issue identification.


Among the three representative instances of rating approaches, ROUGE\_1 and the gte-large-zh embedding model both achieve similarity accuracies of 0.870, but only obtain alignment verification accuracies of 0.635 and 0.665, respectively, with low agreement with human annotations indicated by Cohen's Kappa scores~\cite{cohen1960coefficient}. In contrast, Qwen3.5-27B substantially outperforms the other approaches, achieving 0.965 similarity mapping accuracy, 0.852 alignment verification accuracy, and a Cohen's Kappa of 0.668, indicating substantial agreement with human expert judgment~\cite{landis1977measurement}.


\section{Further Analysis on the LIC Task}

We further analyse legal issue classification (LIC) performance on the subset of model predictions whose issue descriptions are \textit{successfully matched} to reference issues. This analysis isolates legal issue classification performance from issue generation quality and examines whether models can correctly identify the legal nature and doctrinal function of disputed issues once the underlying issues have been correctly identified.

Results show that models frequently fail to accurately determine the legal attributes of issues even when they correctly describe the underlying disputes (Table~\ref{tab:lic_results}). Although all evaluated models achieve strong performance on Tier-1, performance drops substantially as hierarchical depth increases. This decline is particularly pronounced for legal-domain LLMs such as LegalOne-R1 and DISC-LawLLM, while even frontier models such as GPT-5.2 and Gemini-3-Flash exhibit performance gaps of around 0.4 between Tier-1 and Tier-3 settings. These findings suggest that current models still struggle to understand the legal nature and reasoning function of disputed issues, highlighting the importance of explicitly modelling structured legal categories in legal issue identification.

\section{Conclusion}

In this work, we explicitly model legal issue identification as a computational task and introduce a comprehensive resource suite for studying legal issues in litigation, including a legally grounded schema, an operationalised task formulation, an expert-annotated evaluation benchmark, and an expert-curated legal issue knowledge base for retrieval-augmented reasoning. Evaluations on the LexIssue benchmark show that legal issue identification remains challenging for existing large language models, particularly in understanding the legal nature and reasoning roles of disputed issues. Future work may explore more effective approaches for incorporating external legal knowledge, and training-based methods that better align models with litigation-oriented issue identification.

\section*{Limitations}

This work has several limitations. First, the inputs to our legal issue identification task are synthetic claim forms and defences reconstructed from court decisions, rather than original litigation materials submitted by parties. This design reflects a practical constraint: authentic claim forms, defences, and other pleadings are often difficult to obtain at scale. Reconstructing party submissions from judicial decisions therefore provides a useful and reasonably proxy for the information available at the pleadings stage. However, such synthetic inputs cannot fully capture the form, style, strategic framing, and evidential detail of materials produced in real litigation. Future work could therefore explore ways to obtain or construct more authentic claim and defence materials, enabling a closer simulation of legal issue identification as it arises in actual dispute resolution.

Second, evaluating legal issue identification remains inherently challenging. As discussed in Sections~\ref{sec:schema} and~\ref{sec:eval_analysis}, issues in dispute are often expressed in free-form legal language, vary in granularity, and may overlap substantially in meaning without being textually similar. To address this difficulty, we experiment with multiple matching and similarity-comparison approaches and adopt a robust evaluation protocol. Human validation further supports the reliability of the evaluation pipeline, with the similarity assessment step achieving over 96\% accuracy and the alignment verification step achieving over 85\% accuracy. Nonetheless, evaluation remains an open methodological challenge. Future research could develop more verifiable and scalable ways of operationalising legal issue identification, while preserving the legal meaning, granularity, and precision required for this task.

\section*{Ethical Considerations}

The benchmark is constructed from publicly available court judgments obtained from~\citet{cjo2013}. During data construction, personal information contained in the cases was anonymised to protect individual privacy. The benchmark and legal issue knowledge base introduced in this work are intended solely for responsible research use. All expert annotators participating in the annotation process were fairly compensated in accordance with institutional guidelines.


\bibliography{custom}

\appendix

\section{Schema for Legal Issues}\label{app:schema}

The proposed dual-layer schema establishes a structured framework for defining and representing legal issues in disputes through two complementary dimensions. While a descriptive representation preserves the contextual, case-specific narrative of a legal issue, a rigid, structured classification maps these narratives into a hierarchical taxonomy of legal categories. The complete hierarchical architecture of this schema is detailed in Figure \ref{fig:legal-schema-dirtree} below.

\begin{figure}[htbp]
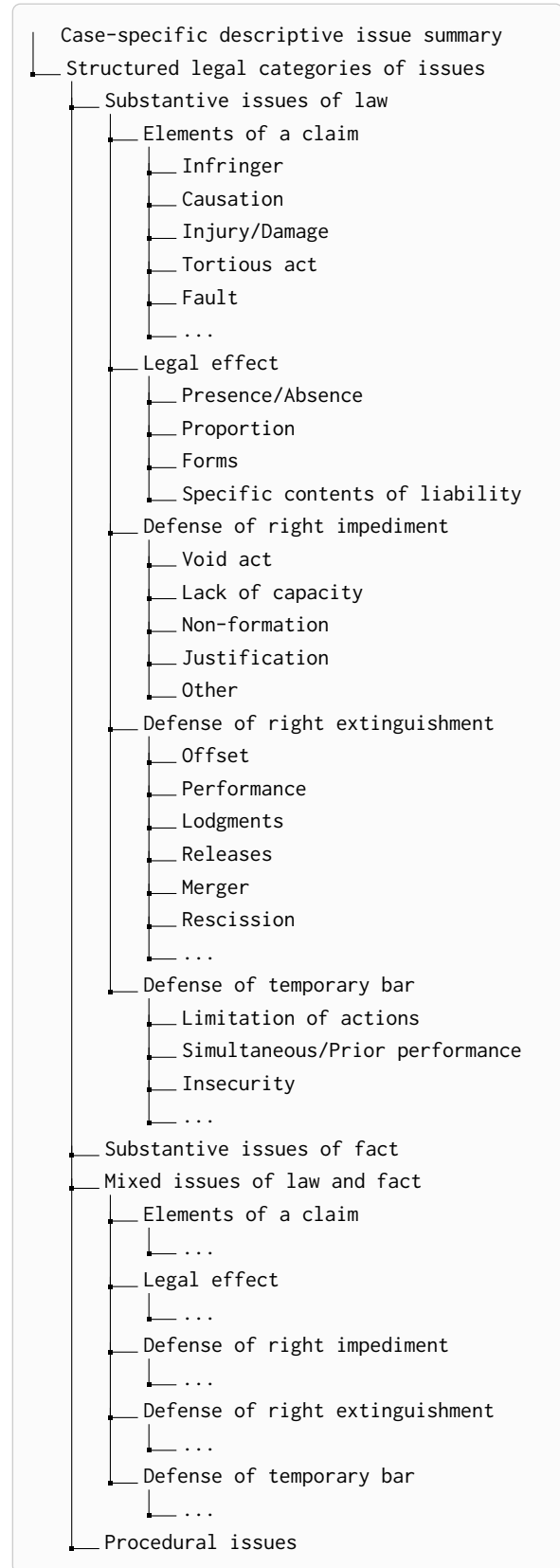

  \centering
  \begin{tcolorbox}[
    colback=gray!3, 
    colframe=gray!40, 
    arc=1mm, 
    boxrule=0.5pt, 
    width=\linewidth,
    left=2mm, right=2mm,
    boxsep=0pt,
    before skip=10pt,
    after skip=10pt
    ]
    \small\ttfamily
    \setlength{\DTbaselineskip}{13pt}
    \dirtree{%
      .1 Case-specific descriptive issue summary.
      .1 Structured legal categories of issues.
      .2 Substantive issues of law.
      .3 Elements of a claim.
      .4 Infringer.
      .4 Causation.
      .4 Injury/Damage.
      .4 Tortious act.
      .4 Fault.
      .4 ....
      .3 Legal effect.
      .4 Presence/Absence.
      .4 Proportion.
      .4 Forms.
      .4 Specific contents of liability.
      .3 Defense of right impediment.
      .4 Void act.
      .4 Lack of capacity.
      .4 Non-formation.
      .4 Justification.
      .4 Other.
      .3 Defense of right extinguishment.
      .4 Offset.
      .4 Performance.
      .4 Lodgments.
      .4 Releases.
      .4 Merger.
      .4 Rescission.
      .4 ....
      .3 Defense of temporary bar.
      .4 Limitation of actions.
      .4 Simultaneous/Prior performance.
      .4 Insecurity.
      .4 ....
      .2 Substantive issues of fact.
      .2 Mixed issues of law and fact.
      .3 Elements of a claim.
      .4 ....
      .3 Legal effect.
      .4 ....
      .3 Defense of right impediment.
      .4 ....
      .3 Defense of right extinguishment.
      .4 ....
      .3 Defense of temporary bar.
      .4 ....
      .2 Procedural issues.
    }
  \end{tcolorbox}
  \caption{The proposed dual-layer hierarchical schema for representing legal issues.}
  \label{fig:legal-schema-dirtree}
\end{figure}

\section{Expert Annotation of Legal Issues}\label{app:human_annotation}

The annotation process for gold-standard issue references involves two legal experts, both holding postgraduate law degrees, under the supervision of a senior legal AI researcher with extensive experience in legal data annotation workflows.

Prior to formal annotation, the annotators received detailed task briefings and jointly reviewed the annotation guidelines to establish a shared understanding of the schema definitions and annotation criteria. To further align annotation standards, the two annotators independently annotated 20 randomly sampled cases and discussed disagreements and ambiguous cases to align annotation standards.

For each case, annotators were provided not only with the reconstructed claim and defence statements, but also with supplementary materials extracted from the original court judgments, including verified case facts, judicial reasoning, and final judgments. These additional materials enabled annotators to more accurately identify the actual disputed legal issues arising in the litigation.

The annotation process requires annotators to provide: (1) descriptive issue summaries, (2) corresponding legal category labels under the schema, and (3) supporting evidence from the original case materials for each annotated issue. Requiring evidence grounding helps ensure that the annotated legal issues are tied to disputes actually discussed in the litigation materials and supported by concrete judicial reasoning or party arguments.

Following the calibration stage, each annotator independently annotated a separate subset of 235 cases. To measure inter-annotator agreement (IAA), 40 cases were deliberately duplicated across annotators. After annotation and deduplication, the resulting dataset contains 430 distinct litigated cases and a total of 1,303 annotated disputed legal issues, each consisting of a descriptive issue summary and corresponding legal category labels. Inter-annotator agreement computed on the 40 overlapping cases achieved a Cohen's Kappa~\cite{cohen1960coefficient} score of 0.687\footnote{According to~\citet{landis1977measurement}, a Cohen's Kappa score between 0.61 and 0.80 signifies ``substantial agreement''.}.

\section{Further Details on Issue Alignment}\label{app:evaluation_design}

The core challenge in evaluating legal issue identification lies in determining whether a predicted issue and a gold-standard issue express the same underlying dispute issue. To address this, we adopt a decomposed prediction-reference alignment pipeline consisting of two stages: \textit{candidate similarity mapping} and \textit{issue-level alignment verification}. 

Formally, for a given case, let the set of dispute issues predicted by the model be: $P = \{p_1, p_2, \dots, p_m\}$, and let the corresponding expert-annotated gold-standard issue set be: $G = \{g_1, g_2, \dots, g_n\}$. Our evaluation framework first establishes prediction-reference alignments $\text{Alignment}(P,G)$, which are obtained through the following steps:



\paragraph{Candidate similarity mapping.} In the first stage, we perform coarse-grained semantic correspondence assessment to retrieve potential issue pairs. For each predicted issue $p_i \in P$, the LLM judge identifies the most semantically similar reference issue $g^*(p_i) \in G$:
$$g^*(p_i) = \arg\max_{g_j \in G} \text{SimJudge}(p_i, g_j)$$

This stage establishes initial candidate pairs $\mathcal{C} = \{(p_i, g^*(p_i)) \mid p_i \in P\}$ based on broad semantic relatedness, without enforcing strict legal equivalence.

\paragraph{Issue-level alignment verification.} In the second stage, we perform stricter issue-level alignment verification. First, the candidate pairs in $\mathcal{C}$ are subjected to a semantic matching assessment which determines whether each candidate pair genuinely expresses the same underlying dispute issue. Unlike the previous stage, which focuses on broad semantic relatedness, this stage evaluates issue-level equivalence under a rubric-based criterion designed by legal experts. This assessment yields a verification score $s_{i}$ for each pair:
$$s_{i} = \text{MatchJudge}(p_i, g^*(p_i))$$

Second, as multiple predicted issues may map to the same reference issue during the similarity mapping stage, we enforce one-to-one alignment through a deduplication procedure that retains only the highest-scoring candidate mapping for each reference issue, producing $\mathcal{C}_{\text{dedup}}$.

A pair is considered successfully aligned if $s_{i} > \tau$, where $\tau$ is a threshold selected on a held-out validation split.

The final prediction-reference alignment is therefore defined as:
$$\text{Alignment}(P,G) = \{(p_i, g_j) \in \mathcal{C}_{\text{dedup}} \mid s_i > \tau \}$$

\section{Prompt Specifications for Issue-Level Alignment Verification}\label{app:rubric_prompt}

\begin{figure*}[t]
  \centering
  \includegraphics[width=\textwidth,height=\textheight,keepaspectratio]{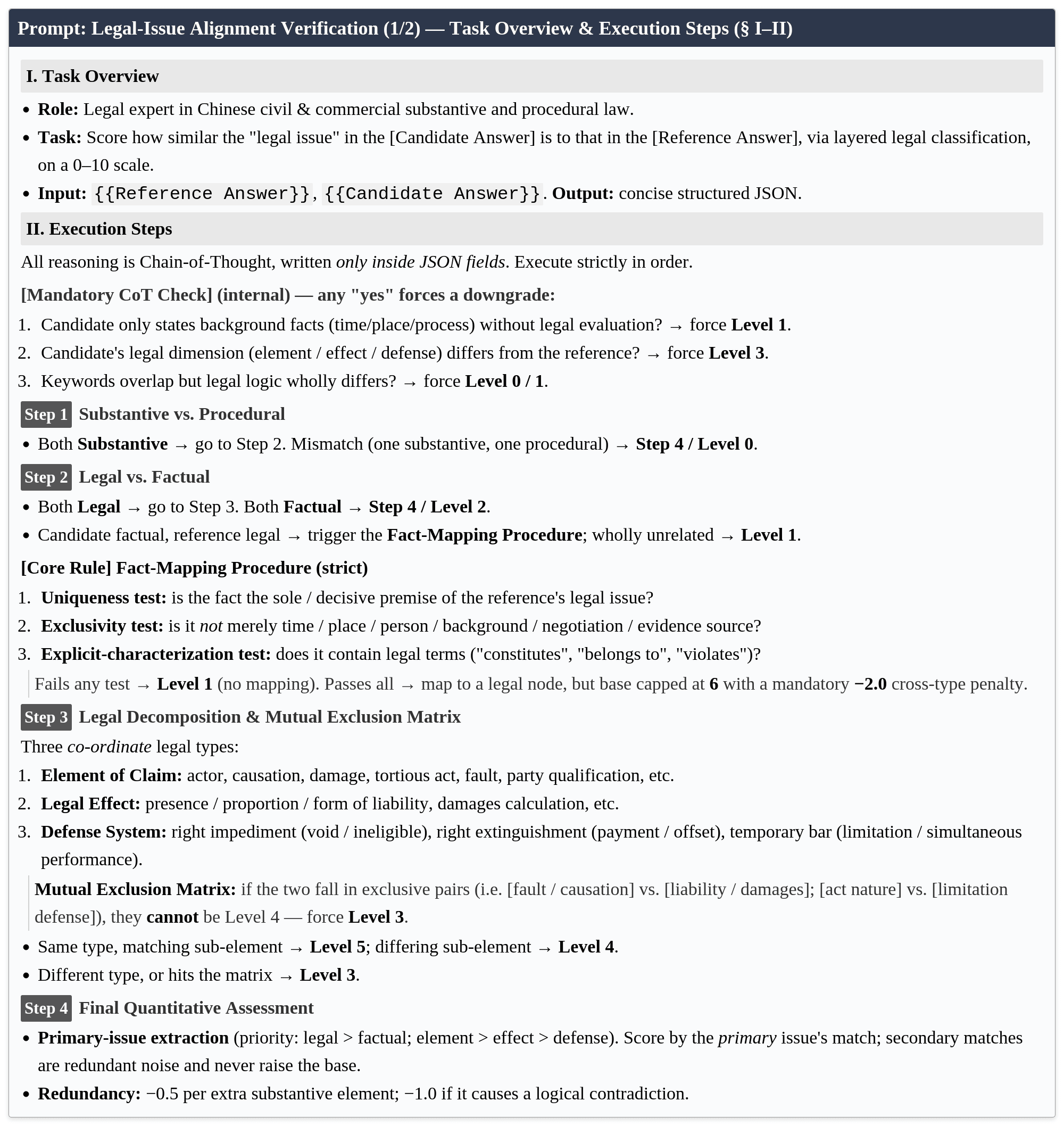}
  \caption{%
    Prompt specification (Part~1/2): Task overview and execution steps (Sections~I \& II).
    Covers the role/task definition, input variables, and the three execution steps
    (Substantive/Procedural $\to$ Legal/Factual $\to$ Legal Logic Subdivision)
    with the Fact-Mapping Procedure for mixed Legal--Factual pairs.
    The Chain-of-Thought analysis is written exclusively in JSON fields.
  }
  \label{fig:legal-prompt-1}
\end{figure*}

\begin{figure*}[t]
  \centering
  \includegraphics[width=\textwidth,height=\textheight,keepaspectratio]{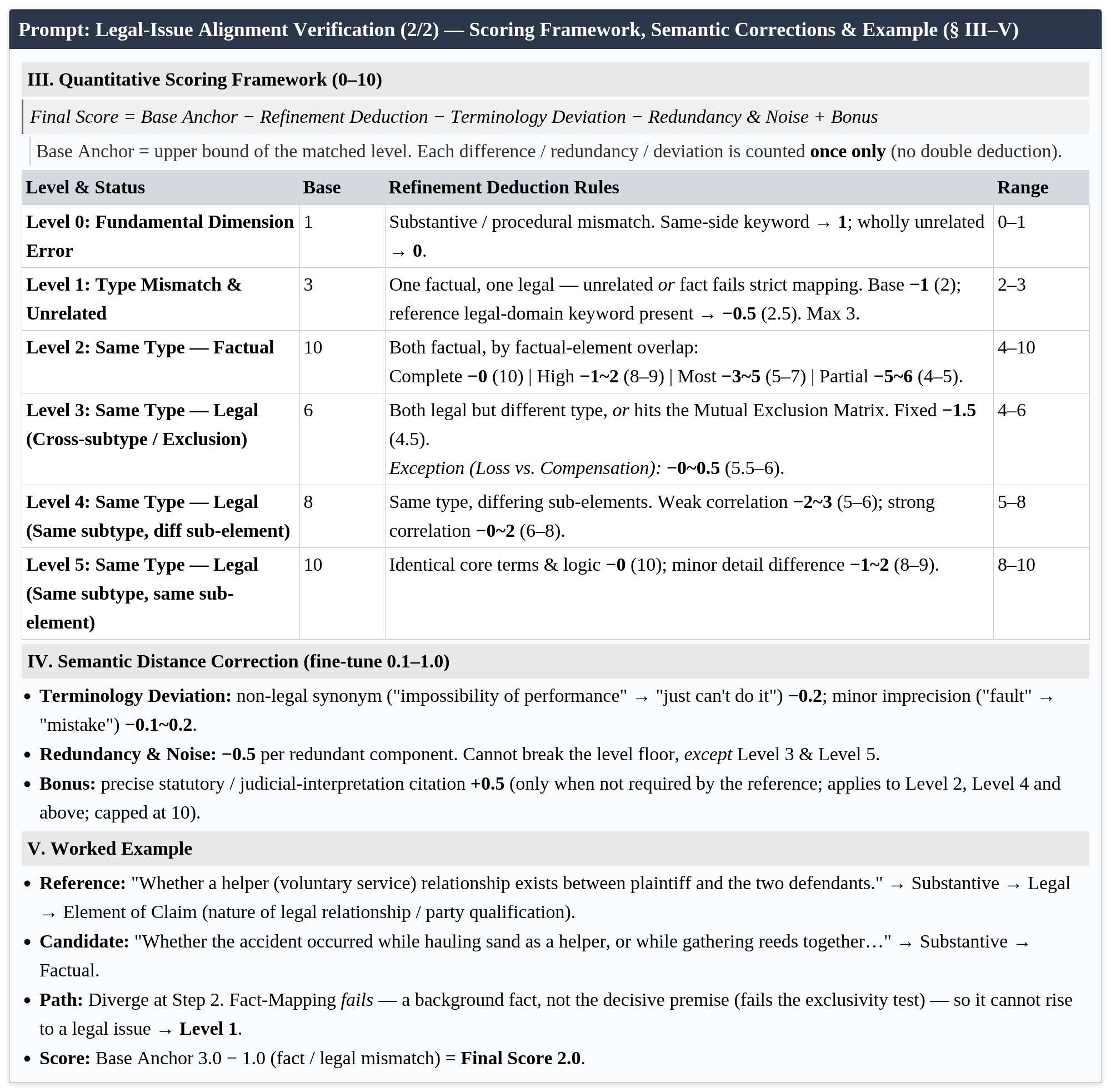}
  \caption{%
    Prompt specification (Part~2/2): Scoring framework, semantic adjustments, and evaluation example (Sections~III, IV \& V).
    Covers the final assessment step (Step~4), the quantitative scoring framework
    with the 6-level benchmark table, semantic distance adjustment rules
    (terminology, redundancy, bonus), and a fully worked example
    with complete classification path and score breakdown.
  }
  \label{fig:legal-prompt-2}
\end{figure*}

\begin{figure*}[t]
  \centering
  \includegraphics[width=\textwidth,height=\textheight,keepaspectratio]{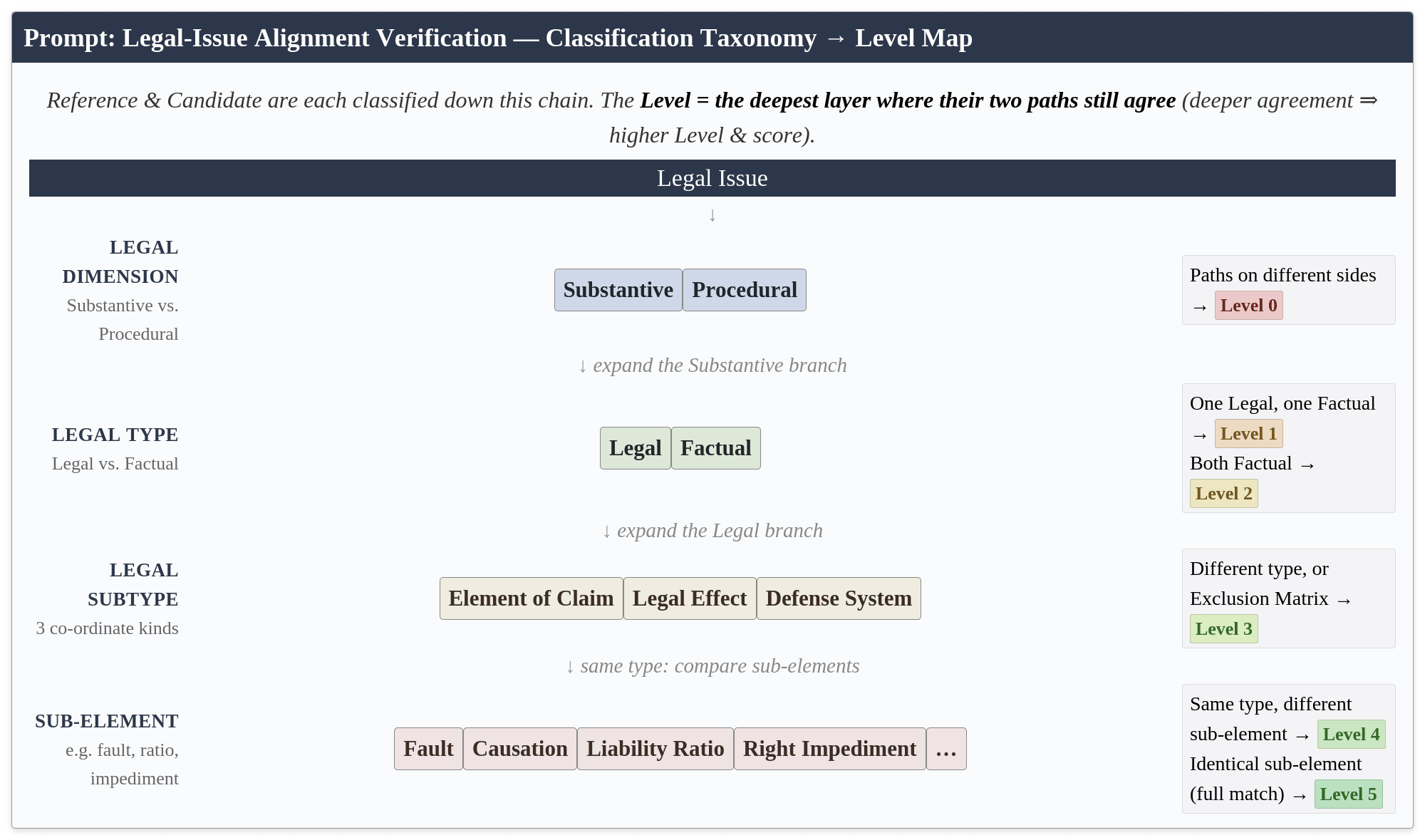}
  \caption{Illustration of the alignment workflow used in the issue-level alignment verification stage. The reference and predicted issues are compared by traversing the hierarchical legal issue taxonomy, with the alignment level determined by the deepest hierarchical tier at which their category paths match. The resulting alignment level forms part of the rubric used by the LLM judge to determine whether the prediction and reference represent the same underlying disputed legal issue.}
  \label{fig:legal-prompt-3}
\end{figure*}

The rubric-based prompt used in the issue-level alignment verification stage is shown in Figures~\ref{fig:legal-prompt-1} (Part1/2) and\ref{fig:legal-prompt-2} (Part2/2). An illustration of the underlying alignment workflow is provided in Figure~\ref{fig:legal-prompt-3}.

\section{Experiment Setup}\label{app:experiment_setup}

For closed-source models, we conduct inference through official API calls using default generation settings to approximate off-the-shelf zero-shot performance. For open-source models, we adopt the recommended inference configurations and default decoding parameters provided in the official model releases, with inference implemented using vLLM.

For the retrieval-augmented generation (RAG) setting, we use Qwen3-Reranker-8B~\cite{qwen3embedding} as the reranking model. The number of retrieved issue candidates is set to $top_k = 5$. 

Inference for open-source models is conducted on a single NVIDIA A800 GPU (80GB). Evaluation scoring is performed with Qwen-3.5-27B~\cite{qwen3.5} using four NVIDIA A800 GPUs (80GB).

\end{document}